\documentclass{article} % For LaTeX2e
\usepackage{iclr2026_conference,times}

\usepackage{amsmath,amsfonts,bm}

\def\eqref#1{equation~\ref{#1}}
\def\1{\bm{1}}

\DeclareMathAlphabet{\mathsfit}{\encodingdefault}{\sfdefault}{m}{sl}
\SetMathAlphabet{\mathsfit}{bold}{\encodingdefault}{\sfdefault}{bx}{n}

\newcommand{\mits}{{\usefont{T1}{ppl}{m}{n}MITS}\xspace}
\newcommand{\mitsfull}{{\usefont{T1}{ppl}{m}{n}MITS-F}\xspace}

\usepackage[T1]{fontenc}
\usepackage[utf8]{inputenc}

\usepackage{hyperref}
\usepackage{url}
\usepackage{latexsym}

\usepackage{graphicx}
\usepackage{subcaption}
\usepackage{enumitem}
\usepackage{booktabs}
\usepackage{float}
\usepackage{xspace}
\usepackage{xcolor}
\usepackage{multirow}
\usepackage{amsmath,amsthm,amssymb,mathtools}
\usepackage{amsfonts}       % blackboard math symbols
\usepackage{wrapfig}

\usepackage{textcomp}
\usepackage{listings}

\makeatletter
\newcommand{\IfCameraDo}[1]{%
  \@ifundefined{iclrfinaltrue}{}{%
    \ificlrfinal #1\fi
  }%
}
\makeatother

\title{\mits: Enhanced Tree Search Reasoning \\ for LLMs via Pointwise Mutual Information}

\author{Jiaxi Li$^1$\footnotemark[1] \quad Yucheng Shi$^1$\thanks{Equal Contribution} 
\quad Xiao Huang$^2$
\quad Jin Lu$^1$\footnotemark[2] \quad Ninghao Liu$^2$\thanks{Corresponding Author}\\
$^1$University of Georgia \quad \quad $^2$The Hong Kong Polytechnic University
}

\iclrfinalcopy % Uncomment for camera-ready version, but NOT for submission.
\begin{document}

\maketitle

\IfCameraDo{\vspace{-1em}}
% \vspace{-1em} % if \iclrfinalcopy

\begin{abstract}
Tree search has become a representative framework for test-time reasoning with large language models (LLMs), exemplified by methods such as Tree-of-Thought and Monte Carlo Tree Search.
However, it remains difficult to provide instant and reliable quantitative assessments of intermediate reasoning step quality, and extensive path exploration is computationally costly.
To address this, we propose Mutual Information Tree Search (\mits), a novel framework that guides reasoning with information-theoretic principles. \mits introduces an effective scoring function based on pointwise mutual information (PMI), which enables step-wise evaluation of reasoning paths and search tree expansion via beam search without expensive look-ahead simulations, achieving superior reasoning performances while maintaining computational efficiency. The framework is complemented by an entropy-based dynamic sampling strategy that adaptively allocates computational resources to uncertain reasoning steps where exploration is most beneficial.
For final prediction, \mits employs a weighted voting scheme that combines PMI scores with prediction consensus.
Through comprehensive experiments on diverse reasoning benchmarks, \mits consistently surpasses baseline methods, establishing a principled and efficient framework for LLM reasoning. The code is available at \url{https://github.com/plusnli/MITS}.
\end{abstract}

\IfCameraDo{\vspace{-1em}}

\section{Introduction}
\label{sec:intro}
Complex multi-step reasoning remains a fundamental challenge for Large Language Models (LLMs), particularly in tasks that require logical deduction, mathematical computation, or systematic problem-solving~\citep{yang2025qwen3, zhu2024multilingual, yi2024survey}. While Chain-of-Thought (CoT) prompting~\citep{wei2022chain, kojima2022large} has emerged as a powerful technique to enhance reasoning by decomposing problems into intermediate steps, it typically generates a single reasoning path, which may lead to incorrect solutions due to error accumulation or the selection of suboptimal reasoning strategies. This limitation becomes particularly pronounced in complex reasoning tasks where multiple valid approaches exist, but only specific paths lead to correct answers.

Recent work on \textit{inference-time scaling law}~\citep{snell2024scaling} reveals a promising direction: generating multiple reasoning paths can substantially improve accuracy. The key insight is that solution coverage, \textit{i.e.}, the probability of finding at least one correct answer, scales predictably with the number of attempts. This observation has been proven valuable in domains with verifiable answers, such as code generation and theorem proving~\citep{wei2025swe}. 
However, exhaustively enumerating reasoning paths is infeasible, as the search space grows exponentially with problem complexity. This leads to the central research problem: \textit{How can we efficiently search through the vast space of possible reasoning paths to identify those most likely to yield correct solutions?}

To tackle this problem, existing approaches face several limitations. Methods based on exhaustive rollouts or Monte Carlo Tree Search (MCTS;~\citealp{rap, qi2025mutual, ding-etal-2024-everything}) require extensive forward simulation, which is computationally prohibitive at scale. 
Self-evaluation methods struggle to provide quantitative assessments of reasoning quality, often relying on pairwise comparisons or binary judgments that fail to capture nuanced differences between paths~\citep{xie2023self, gu2024survey}. 
Most critically, existing scoring mechanisms usually favor \textbf{plausible but generic reasoning paths that could apply to many problems, rather than those specifically tailored to the given question}~\citep{sharma2024towards}.
These limitations highlight the need for a principled framework that can (i) evaluate reasoning quality efficiently without costly simulations, and (ii) distinguish between generic reasoning and question-specific reasoning.

In this work, we propose Mutual Information Tree Search (\mits), a principled framework that addresses these challenges through information-theoretic guidance. Our key insight is that effective reasoning paths should exhibit high mutual information with the question, which means they should contain information that is both relevant and uniquely tied to solving the given problem. 
We leverage Pointwise Mutual Information (PMI) as a scoring mechanism that quantifies how much a reasoning path's plausibility increases \emph{because of the specific question}, effectively filtering out generic or spurious reasoning patterns. 
Moreover, by computing scores dynamically as the reasoning process unfolds, \mits evaluates reasoning quality efficiently without expensive look-ahead simulations.

Overall, our framework introduces three technical innovations. 
First, we propose a \textbf{principled scoring approach} for intermediate reasoning steps using Pointwise Mutual Information (PMI), which reduces computational burden and establishes reliable criteria for trajectory selection.
Second, we propose an \textbf{entropy-based dynamic sampling strategy} that adaptively allocates computational resources according to the uncertainty at each reasoning step, concentrating exploration on uncertain decision points where diverse reasoning approaches yield the greatest potential benefit.
Third, we introduce a \textbf{weighted voting scheme} that combines PMI scores with prediction consensus, reducing the risk of selecting high-scoring but spurious reasoning paths by utilizing agreement across multiple reasoning trajectories.
Through extensive experiments across diverse reasoning benchmarks, we show that \mits achieves substantial improvements over several strong baseline methods. 

The remainder of this paper is organized as follows: Section~\ref{sec:prelim} reviews background on tree search methods for LLM reasoning and current challenges, Section~\ref{sec:method} presents our \mits framework in detail, Section~\ref{sec:exp} evaluates our approach on multiple reasoning benchmarks, Section~\ref{sec:related} summarizes related work, and Section~\ref{sec:conclusion} gives conclusion, discusses broader implications and future directions.

\IfCameraDo{\vspace{-0.5em}}
\section{Preliminary}
\label{sec:prelim}

\paragraph{Tree-Search Based Reasoning.}
To address reasoning tasks with large language models (LLMs), the problem is often framed as a multi-step generation process with chain-of-thought (CoT) prompting~\citep{wei2022chain, kojima2022large}, where the model decomposes and solves the problem step by step.
However, CoT is restricted to a single reasoning trajectory, which limits exploration of the solution space.
To overcome the limitation, more and more works adopt tree-based frameworks~\citep{yao2023tree, rap, qi2025mutual} that enable systematic exploration of multiple reasoning paths.
Specifically, given a reasoning problem \(q\), we construct a search tree \(\mathcal{T}\) by producing and sampling the nodes step by step, where the root node corresponds to the question \(q\) and each child node is an intermediate reasoning step \(s\). A path from the root node \(q\) to a node \(s_n (n \ge 1)\) constitutes a candidate solution path \(S=q\oplus s_1\oplus s_2\oplus \dots \oplus s_n\), where the operator \(\oplus\) denotes the sequential concatenation. 
Within the search tree \(\mathcal{T}\), a set of solution paths \(\mathbb{S}=\{S^1, S^2, \dots, S^c\} (c\ge1)\) is extracted, from which we aim to find the most plausible path as the final answer.

\IfCameraDo{\vspace{-0.5em}}

\paragraph{Challenges in Tree Search.}
Recently, within the tree-search paradigm~\citep{yao2023tree} for LLM reasoning, Monte-Carlo Tree Search (MCTS) emerges as one of the most powerful and widely-used algorithms~\citep{rap, qi2025mutual, ding-etal-2024-everything}. 
However, evaluating the quality of intermediate reasoning steps in MCTS typically relies on repeated look-ahead rollouts to simulate future trajectories and back-propagate for value estimation, incurring substantial computational cost. 
In addition, prior work highlights the difficulty of defining a reliable metric for selecting the correct trajectory~\citep{qi2025mutual}. 
To this end, we ask whether we can design a principled scoring function that can assess the quality of reasoning traces without simulated rollouts, thereby addressing both the computational efficiency and trajectory selection challenges simultaneously.

\begin{figure}
    \centering
    \includegraphics[width=0.96\linewidth]{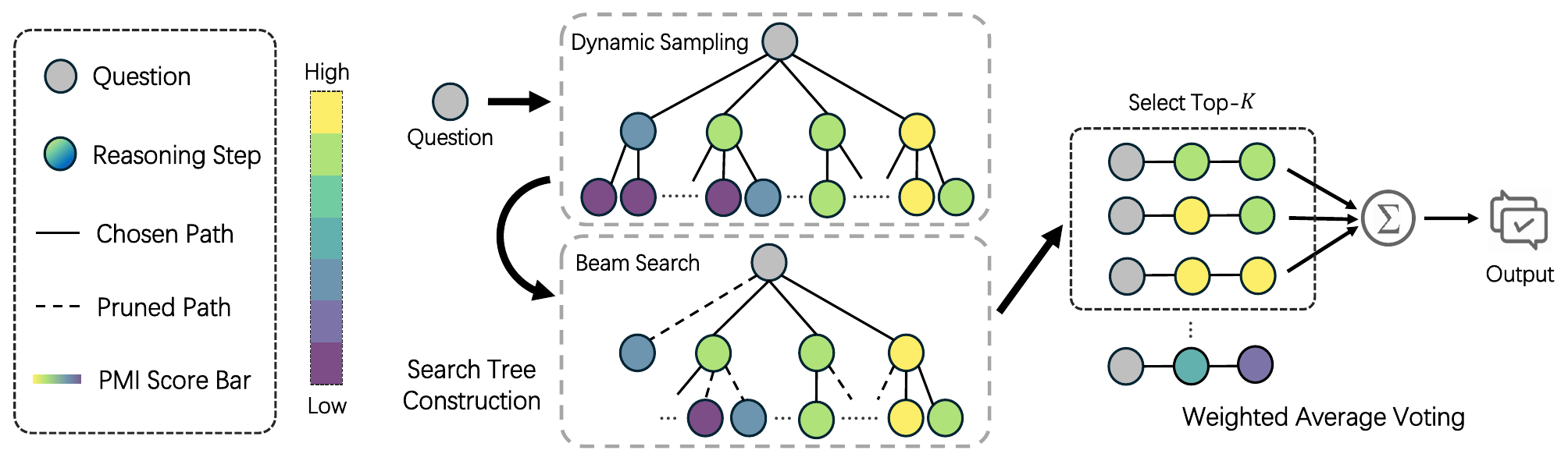}
    \caption{Overview for the \mits approach. The search tree is incrementally constructed by generating reasoning steps conditioned on prior context. Dynamic sampling controls the number of candidates (i.e., nodes) generated at each step. Beam search is applied to prune less promising paths. Weighted average voting is used to aggregate candidates and select the final output.}
    \label{fig:overview}
    \vspace{-5pt}
\end{figure}

\section{Methodology}
\label{sec:method}

This section introduces the proposed Mutual Information Tree Search (\mits) method, which constructs a search tree and evaluates reasoning path quality through an effective mutual information based criterion, without relying on looking-ahead simulations. As illustrated in Figure~\ref{fig:overview}, there are three major components as follows.

\textbf{1. Pointwise Mutual Information (PMI) Scoring.} 
Given a question and its reasoning path, our objective is to evaluate the path’s quality. 
We employ Pointwise Mutual Information (PMI) as a principled and effective scoring metric for intermediate reasoning steps (section~\ref{sec:pmi}), which then serves as the criterion for guiding search tree expansion.

\textbf{2. Search Tree Construction.} Using PMI as guidance, we build the search tree with the beam search technique as demonstrated in section~\ref{sec:tree}. Furthermore, we develop a dynamic sampling strategy, which allocates computation resources dynamically according to the uncertainty of the current step.

\textbf{3. Weighted Average Voting for Output Selection.} The search tree yields a set of candidate reasoning chains, from which we make the final prediction decision as in section~\ref{sec:weight_ave}. 
Unlike standard majority voting~\citep{wang2023selfconsistency}, which treats all predictions equally and overlooks their relative strengths, we propose a weighted average voting scheme that assigns weights to predictions according to their PMI scores, leading to more reliable answer selection.

\subsection{Pointwise Mutual Information (PMI) Scoring}
\label{sec:pmi}

An effective scoring function for reasoning tasks should not only favor solutions that appear plausible, but also quantify how much \textit{specific} information a solution contributes to answering the question~\citep{shi2024retrieval, cui2025entropy, agarwal2025unreasonable}. In particular, the function should:
(i) assign higher scores to solutions that are tightly coupled with the question, and
(ii) penalize generic or spurious solutions that could apply broadly to unrelated queries.
Mutual Information (MI) naturally satisfies these requirements, as it measures the reduction in uncertainty about solutions \(S\) once questions \(q\) are observed. Formally, the mutual information between questions and solutions is:
\begin{equation}
I(S;q) = \mathbb{E}_{p(S,q)} \left[\log \frac{p(S, q)}{p(S)\,p(q)}\right],
\end{equation}
where \(p(S, q)\) is the joint probability of \(q\) and \(S\); \(p(q)\) and \(p(S)\) are the marginal probability of the question and reasoning path, respectively.
While MI captures the average dependence between questions and reasoning paths, the quantity inside the expectation, \(\log \tfrac{p(S, q)}{p(S)\,p(q)}\), evaluates the contribution of a particular question-reasoning pair. This term is known as the \textit{pointwise mutual information (PMI)}.
Since \(\tfrac{p(S, q)}{p(q)}=p(S \mid q)\), the PMI score simplifies to
\begin{equation}
    \label{eq:pmi_def}
    \mathrm{PMI}(q; S) = \log \frac{p(S \mid q)}{p(S)}.
\end{equation}
Here, $\mathrm{PMI}(q;S)$ satisfies the aforementioned requirements for an effective scoring function, because (i) the probability $p(S \mid q)$ measures the degree of match for the reasoning process $S$ given the question $q$, and (ii) the denominator $p(S)$ penalizes generic paths that remain likely even without observing $q$.

\paragraph{Probability Estimation.}
To compute the PMI score, we need to estimate the probabilities \(p(S \mid q)\) and \(p(S)\) in Equation~\ref{eq:pmi_def}.
Suppose a reasoning path \(S\) is currently expanded up to the \(n\)-th step (\(n > 1\)), comprising a sequence of reasoning steps \((s_1, s_2, \dots, s_n)\). The conditional probability \(p(S \mid q)\) is decomposed as \(p(S \mid q) = \prod_{i=1}^{n} p(s_i \mid q, s_1, \dots, s_{i-1})\) and the marginal probability is \(p(S) = \prod_{i=1}^{n} p(s_i \mid s_1, \dots, s_{i-1})\). 
To estimate the conditional probabilities \(p(s_i \mid q, s_1, \dots, s_{i-1})\) and \(p(s_i \mid s_1, \dots, s_{i-1})\), we leverage the next-token prediction abilities of auto-regressive LLMs.
Specifically, both probabilities are computed by multiplying token distributions within step \(s_i\), with \(p(s_i \mid q, s_1, \dots, s_{i-1})\) conditioned on the full context including question \(q\), and \(p(s_i \mid s_1, \dots, s_{i-1})\) conditioned only on previous reasoning steps.
When \(n=1\), \(p(S \mid q) = p(s_1 \mid q)\), and \(p(S) = p(s_1)\). In implementation, to approximate \(p(s_1)\), we prepend the special Begin-of-Sequence (\texttt{<bos>}) token to the language model.

\paragraph{PMI Scoring with Incremental Update.}
\label{sec:incre_upd}
We are now able to calculate the PMI score given a complete question and reasoning chain. However, to extend the reasoning process step-by-step in a search tree, it is preferable to use an incremental update formula rather than re-computing PMI from the beginning at each step.
Let $S_n = (s_1, s_2, \ldots, s_n)$ denote the reasoning path at step $n$, and \(\mathrm{PMI}_n\) be the PMI score up to the $n$-th step.
We can have
\begin{align}
    \label{eq:pmi_n}
    \mathrm{PMI}_n := \mathrm{PMI}(q; S_n) = \sum_{i=1}^{n} \big[ \log p(s_i \mid q, s_1, \ldots, s_{i-1}) - \log p(s_i \mid s_1, \ldots, s_{i-1}) \big].
\end{align}
The detailed derivation can be found in Appendix~\ref{app:pmi_derivation}. We transform it into an additive form using log probabilities for ease of computation.
Then, the incremental update for extending the reasoning path from step $n$ to step $n+1$ is given by:
\begin{equation}
    \label{eq:pmi_recur}
    \mathrm{PMI}_{n+1} = \mathrm{PMI}_n + \big[ \log p(s_{n+1} \mid q, s_1, \ldots, s_n) - \log p(s_{n+1} \mid s_1, \ldots, s_n) \big],
\end{equation}
with the initial condition \(\mathrm{PMI}_0 = 0\), i.e., the PMI score is zero before any reasoning step is generated.
In this way, each update captures the incremental information gain contributed by the new step $s_{n+1}$ toward answering the question $q$, beyond what has already been accounted for by the preceding steps. This not only enables efficient computation during reasoning, but also provides an interpretable view of how individual steps contribute to problem-solving.

\subsection{Search Tree Construction}
\label{sec:tree}

In this subsection, we introduce how to build the reasoning search tree \(\mathcal{T}\) using the PMI criterion. We use a generator model \(G\) to produce each step and an evaluator model \(E\) to compute PMI scores.
We first introduce how to generate each reasoning step, then the construction of the reasoning tree \(\mathcal{T}\), which couples two mechanisms: (1) dynamic sampling that decides how many next-step candidates to propose; (2) PMI-guided beam search that evaluates and prunes candidates to retain only the most promising paths for further expansion. In short, dynamic sampling produces candidates, and beam search selects which of them advance to the next level of the tree.

\begin{wrapfigure}{r}{0.62\textwidth}
  \centering
  \vspace{-1.5em}
  \includegraphics[width=\linewidth]{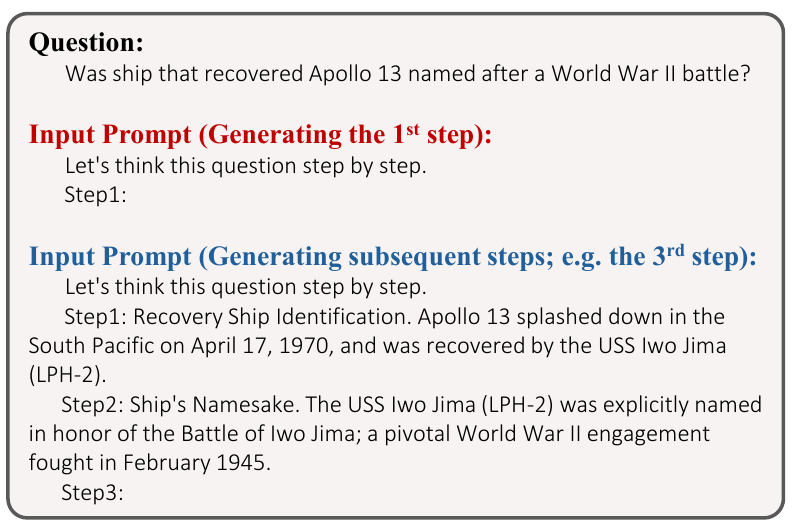}
  \caption{Prompt examples for step-by-step reasoning generation, showing the iterative generation from initial step to subsequent steps, similar to~\citep{lai2024step}.}
  \label{fig:prompt_step}
  \vspace{-1em}
\end{wrapfigure}

\paragraph{Reasoning Step Generation.}
To construct the tree, a fundamental operation is to generate intermediate reasoning steps. At step $n+1$, the generator model $G$ produces \emph{multiple candidate steps} conditioned on the question $q$ and the partial reasoning path $(s_1,\ldots,s_{n})$. 
As shown in Figure~\ref{fig:prompt_step}, we format generation with a CoT-style instruction (\textit{e.g.}, "Let’s think step by step") and delimit each step using a special marker (\textit{e.g.}, \texttt{"Step"}), so that $G$ produces one step at a time before returning control for evaluation by evaluator model \(E\). 
The first set of candidates is conditioned solely on $q$, while later candidates are conditioned on both $q$ and the preceding reasoning context.

\paragraph{Dynamic Sampling via Entropy Quantification.}
Through reasoning step generation, we can produce multiple candidate steps conditioned on the previous one. However, how many candidate steps to sample requires careful consideration to optimize test-time compute~\citep{singhi2025solve, snell2024scaling}. In particular, more computational resources should be allocated to reasoning branches with higher \textit{uncertainty}, since they offer greater potential for generating and exploring diverse paths.

To this end, we use entropy to quantify uncertainty and sample next-step candidates dynamically.
For a reasoning step $s_i$, its entropy is computed over the token distribution as:
\begin{equation}
  \label{eq:entropy}
  H_i = -\textstyle{\sum}_{v \in \mathcal{V}} \,\, p_i(v) \log p_i(v),
\end{equation}
where \(\mathcal{V}\) is the vocabulary and \(p_i(v)\) denotes the probability of the token \(v\) at step $i$. 
To adjust sampling adaptively, we divide the entropy into three intervals: high, moderate, and low.
However, it is hard to use fixed thresholds to divide them, because entropy values can fluctuate given different tasks or different questions due to varying difficulty. 
To address this, we maintain a history of entropies for all generated steps \(H_{1:m} = \{H_1, H_2, ..., H_m\}\), where \(m\) is the total number of nodes in the current search tree, and use empirical \textit{quantiles} to compute adaptive thresholds.
Specifically, we set \(H_\text{low} = \text{percentile}(H_{1:m}, 33\%)\) and \(H_\text{high} = \text{percentile}(H_{1:m}, 67\%)\), which automatically partition entropy values into three regions: low ($H_i < H_\text{low}$), moderate ($H_\text{low} \leq H_i \leq H_\text{high}$), and high ($H_i > H_\text{high}$). Unlike fixed cutoffs, these quantile-based thresholds adapt to the varying difficulty across tasks and questions.
Finally, the number of samples for step $i+1$ is adjusted based on the entropy region. The final sampling number is set to \(N_\text{base} + \Delta N_i\), where \(N_\text{base}\) is the default sample size and \(\Delta N_i\) is determined by the entropy level. The complete formulation for computing \(\Delta N_i\) is given in Appendix~\ref{app:dynamic_samp}.

\paragraph{Beam Search with PMI.}
Once candidate steps are generated, the evaluator model $E$ computes their incremental PMI contributions, and updates the cumulative score to $\mathrm{PMI}_{n+1}$ according to Equation~\ref{eq:pmi_recur}. 
Since generating and evaluating all possible paths quickly becomes computationally expensive, we employ \textit{beam search pruning} to ensure efficiency. Candidates are ranked by their cumulative PMI scores, and only the top-$B$ paths are retained for the next step, where \(B\) denotes the beam width. 
This pruning keeps the tree computationally tractable while focusing exploration on the most informative reasoning directions. 
By tightly coupling candidate generation with PMI-based pruning, the method balances diversity and quality throughout the tree search.

\subsection{Weighted Average Voting for Output Selection}
\label{sec:weight_ave}
After building the search tree \(\mathcal{T}\), we need to select the final output from the collected reasoning chains \(\mathbb{S} = \{S^1, S^2, \ldots, S^c\}\), where \(c\) is the total number of candidate chains. 
A straightforward approach is to choose the chain with the highest PMI score. However, this can be brittle: the top-scoring chain may arise from spurious correlations between the query and particular reasoning steps~\citep{zhao2024explainability, shao2025spurious, xie2025mitigating}, or from overfitting to specific linguistic patterns~\citep{wu2025self, li2025fact}, leading to overconfident yet incorrect predictions.
To mitigate this risk, we propose Weighted Average Voting, a simple but effective post-hoc operation by incorporating \textit{prediction frequency} as a weight on PMI scores, as consensus among multiple reasoning paths often indicates more reliable conclusions. 

Formally, given the reasoning chains, we first sort them according to PMI scores from high to low, and then select the top \(K\) chains: \(\mathbb{S}_{\text{top-}K} = \{S^{1}, S^{2}, \ldots, S^{K}\}\). For each chain \(S \in \mathbb{S}_{\text{top-}K}\), we extract its final prediction \(\mathrm{Pred}(S)\). The frequency of each unique prediction is then computed as: \(\mathrm{Freq}(p) = |\{S \in \mathbb{S}_{\text{top-}K} : \mathrm{Pred}(S) = p\}|\).
After that, we can compute the frequency-weighted PMI score (\(\mathrm{PMI}^{*}\)) for each reasoning chain:
\begin{equation}
    \label{eq:weighted_pmi}
    \mathrm{PMI}^{*}(q;S) = \mathrm{PMI}(q; S) \cdot \frac{\mathrm{Freq}(\mathrm{Pred}(S))}{K}.
\end{equation}
We re-rank the top \(K\) chains according to \(\mathrm{PMI}^{*}(q; S)\) and select the chain with the highest one as our final prediction: \(S^* = \arg\max_{S \in \mathbb{S}_{\text{top-}K}} \mathrm{PMI}^{*}(q; S).\)
This reweighting mechanism provides a regularization effect and achieves a balance between confidence and consensus.

\section{Experiments}
\label{sec:exp}

\subsection{Setup}
\label{sec:exp_setup}

\paragraph{Backbone models and datasets.}
\mits is a general algorithm and applicable to various LLMs and different reasoning tasks. 
We conduct main experiments on the LLMs \textsc{Qwen2.5-3B-Instruct}, \textsc{Qwen2.5-7B-Instruct}~\citep{qwen2.5}, \textsc{Phi-3.5-mini-Instruct} and \textsc{Phi-4-mini-Instruct}~\citep{abdin2024phi}. All used models are instruction-tuned, and we omit the \texttt{-Instruct} suffix for brevity in the following sections and tables.
We perform experiments on three datasets, StrategyQA~\citep{geva-etal-2021-aristotle} of questions that require implicit multi-hop strategies to answer, ARC-Challenge~\citep{clark-etal-2018-arc}, which represents a scientific knowledge-intensive reasoning task, and CommonsenseQA~\citep{talmor-etal-2019-commonsenseqa} of commonsense questions that require prior world knowledge to answer.

\paragraph{Implementation details.}
When calculating the PMI score~(\ref{eq:pmi_n}), we observe that a single-step PMI value tends to be positive, \textit{i.e.}, \(p(s_{n+1} \mid q, s_1, \ldots, s_n) > p(s_{n+1} \mid s_1, \ldots, s_n)\), which induces a length bias whereby longer reasoning chains tend to yield larger PMI scores. To mitigate this effect, we normalize PMI through multiplying \(\mathrm{PMI}_n\) by factor \(\frac{1}{n}\). % average and sum
In the stage of search tree construction, the maximum depth of the tree is set to 10, and the default sampling number \(N_\text{base}=3\). The evaluator model \(E\) is set as same with the generator model \(G\) by default.
We set the beam width \(B\) at 32 for the beam search process of \mits. 
We also run \mitsfull, the variant algorithm of \mits by fully expanding the search tree without beam search, which allows us to analyze the trade-off between computational efficiency and reasoning performance.
We set \(K=32\) for weighted majority voting for both of \mits and \mitsfull.
To guarantee reproducible inference process, we searched a set of hyperparameters, which is detailed in Appendix~\ref{app:hyperparams}.

\begin{table*}[!t]
  \centering
  \small
  \caption{Accuracies of \mits compared with several baselines on different datasets. Columns are grouped by LLM backbones. Best results are in \textbf{bold}, and second best results are \underline{underlined}.}
  \label{tab:results}
  \setlength{\tabcolsep}{8pt}
  \begin{tabular}{lcccc}
    \toprule[1.6pt]
    \textbf{Method} & \textbf{\textsc{Qwen2.5-3B}} & \textbf{\textsc{Qwen2.5-7B}} & \textbf{\textsc{Phi-3.5-mini}} & \textbf{\textsc{Phi-4-mini}} \\
    \midrule[1.3pt]
    \multicolumn{5}{c}{\textit{StrategyQA}} \\
    CoT           & 47.34 & 65.47 & 54.47 & 51.25 \\
    CoT-SC    & 56.12 & 68.74 & 56.71 & 55.47 \\
    ToT           & 49.86 & 66.54 & 54.12 & 57.41 \\
    RAP           & 60.56 & 68.14 & 57.36 & 59.45 \\
    rStar         & 65.32 & 70.41 & 62.57 & 64.89 \\
    \midrule[1pt]
    \mitsfull       & \underline{67.84} & \underline{74.73} & \underline{65.75} & \textbf{70.45} \\
    \mits   & \textbf{68.45} & \textbf{75.76} & \textbf{66.47} & \underline{68.19} \\
    \midrule[1.3pt]
    \multicolumn{5}{c}{\textit{ARC-Challenge}} \\
    CoT           & 69.11 & 71.50 & 67.52 & 72.25 \\
    CoT-SC    & 81.16 & 81.02 & 77.95 & 82.48 \\
    ToT           & 85.45 & 85.86 & 77.86 & 83.56 \\
    RAP           & 83.96 & 83.45 & 83.56 & 85.48 \\
    rStar         & 86.74 & 87.24 & 86.84 & 87.74 \\
    \midrule[1pt]
    \mitsfull       & \underline{87.45} & \textbf{93.45} & \underline{90.54} & \textbf{91.56} \\
    \mits   & \textbf{90.68} & \underline{92.55} & \textbf{90.85} & \underline{90.35} \\
    \midrule[1.3pt]
    \multicolumn{5}{c}{\textit{CommonsenseQA}} \\
    CoT           & 66.67 & 66.50 & 55.86 & 65.71 \\
    CoT-SC    & 74.72 & 76.52 & 67.93 & 69.82 \\
    ToT           & 75.56 & 79.32 & 66.49 & 72.56 \\
    RAP           & 73.24 & 82.23 & 73.23 & 75.84 \\
    rStar         & \underline{79.34} & \underline{85.73} & 77.65 & 79.41 \\
    \midrule[1pt]
    \mitsfull       & \textbf{79.68} & \textbf{86.25} & \textbf{79.56} & \textbf{81.49} \\
    \mits   & 78.83 & 84.80 & \underline{78.14} & \underline{80.28} \\
    \bottomrule[1.6 pt]
  \end{tabular}
\end{table*}

\subsection{Main Results}

\paragraph{Baselines.} We compare \mits with three types of baselines: single CoT prompting, multiple CoT sampling and tree-search methods. Since \mits only leverages inference phase of language models, we focus on training-free baselines for fair comparison.
\textbf{(1)} For single CoT prompting, we use CoT prompting~\citep{wei2022chain, kojima2022large}. % zero-shot
\textbf{(2)} For multiple CoT sampling, we choose the widely adopted CoT with self-consistency method (CoT-SC;~\citealp{wang2023selfconsistency}), and employ majority voting to select the final prediction. We report the performances of sampling 32 times.
\textbf{(3)} For tree-search methods, we include three strong baseline methods, Tree-of-thought (ToT;~\citealp{yao2023tree}), RAP~\citep{rap} and rStar~\citep{qi2025mutual}. We run ToT with Breath-First Search (BFS) to expand the tree. RAP and rStar utilize Monte-Carlo Tree Search algorithm (MCTS;~\citealp{utc}) for tree search process, with 32 rollouts performed for each method.

\paragraph{Performances on diverse reasoning datasets.}
Here we present the performances of \mits against several baselines on diverse reasoning datasets in Table~\ref{tab:results}. We start by evaluating the effectiveness of \mits on general reasoning benchmarks.
We highlight three observations. 
\textbf{(1)} \mits demonstrates substantial improvements over previous approaches across diverse tasks. For example, on StrategyQA dataset with \textsc{Qwen2.5-3B}, \mits achieves 68.45\% accuracy compared to CoT's 47.34\%, representing a remarkable +21.11\% improvement. Similarly, on ARC-Challenge with \textsc{Qwen2.5-7B}, \mits reaches 92.55\% accuracy, significantly outperforming the strongest baseline rStar by +5.31\%. This substantial gain indicates that PMI-based scoring provides more effective guidance for reasoning path selection.
\textbf{(2)} \mits consistently outperforms existing methods across different model scales and reasoning domains. Unlike baseline methods that exhibit varying effectiveness across different tasks, \mits maintains strong results across logical reasoning (StrategyQA), scientific reasoning (ARC-Challenge), and commonsense reasoning (CommonsenseQA). For instance, while RAP achieves competitive results on some tasks, it falls significantly behind \mits on StrategyQA (60.56\% vs 68.45\% on \textsc{Qwen2.5-3B}). 
\textbf{(3)} The comparison between \mits and \mitsfull reveals the value of beam search in reasoning. While \mitsfull shows improvements on CommonsenseQA, it generally underperforms \mits on StrategyQA (e.g., 74.73\% vs 75.76\% on \textsc{Qwen2.5-7B}). This indicates that PMI-based beam search can enhance performance by filtering out inferior reasoning paths early, preventing exploration from being derailed by low-quality branches.

\paragraph{Computational efficiency comparison.}
To validate the computational efficiency of \mits, we conduct experiments on the StrategyQA dataset using \textsc{Qwen2.5-3B}, comparing against CoT-SC, RAP, and rStar under identical inference settings. We measure wall-clock time from inference start to reasoning completion and report the average time that each method needs to solve a problem.

\begin{wraptable}[9]{r}{0.42\linewidth}
  \vspace{-8pt}
  \centering
  \small
  \captionsetup{width=\linewidth,justification=centering}
  \caption{Compute efficiency comparison.}
  \label{tab:efficiency}

  \setlength{\tabcolsep}{4pt}
  \renewcommand{\arraystretch}{1.1}

  \begin{tabular}{lccc}
    \toprule
    \textbf{Method} & \textbf{Time (s)} & \textbf{Acc. (\%)} \\
    \midrule
    CoT-SC & 2.75   & 56.12  \\
    RAP    & 203.42 & 60.56  \\
    rStar  & 815.67 & 65.32  \\
    \mits  & 64.41  & 68.45  \\ 
    \bottomrule
  \end{tabular}
\end{wraptable}

Table~\ref{tab:efficiency} presents the results, revealing significant efficiency advantages for \mits. While CoT-SC achieves the fastest time (2.75s), it delivers the lowest accuracy (56.12\%). MCTS-based methods RAP and rStar require substantially more computation time (203.42s and 815.67s respectively, which are 3.2× and 12.7× slower than \mits) with limited accuracy improvements (60.56\% and 65.32\%). In contrast, \mits achieves the highest accuracy (68.45\%) with moderate computational cost (64.41s), demonstrating the optimal accuracy-per-time trade-off among all methods. These results validate our claim that PMI-based scoring achieves superior reasoning performance with significantly reduced computational overhead compared to previous methods.

\subsection{Ablation Study}

\begin{figure}[t]
    \centering
    \includegraphics[width=\textwidth]{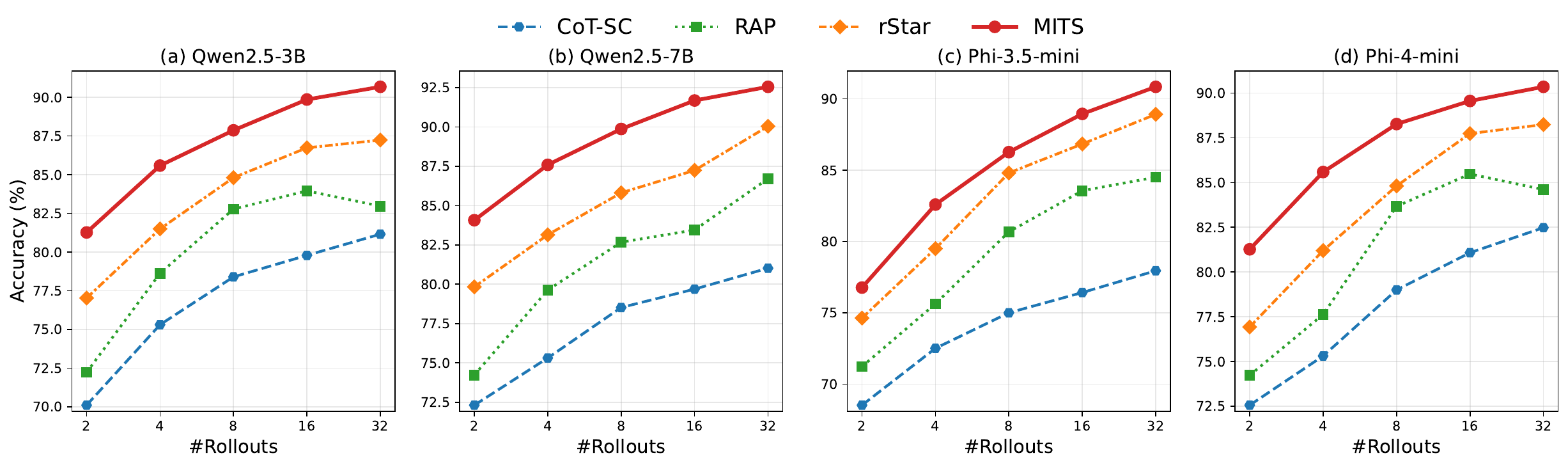}
    \caption{Performance comparison across different rollouts on the ARC-Challenge dataset. \mits not only demonstrates superior performance in early stages with fewer rollouts, but also continues to improve with increased computational budget.}
    \label{fig:perf_rollouts}
\end{figure}

\paragraph{Effectiveness under different number of rollouts.}
MCTS-based baseline methods such as RAP and rStar construct the search tree through a rollout policy of MCTS algorithm. Increasing the number of rollouts expands the set of candidate solution trajectories and can lead to performance improvements, but it also incurs higher inference costs. 
Here we compare the accuracies of CoT-SC, RAP, rStar and \mits under different number of rollouts on the ARC-Challenge dataset as shown in Figure~\ref{fig:perf_rollouts}.
For CoT-SC and \mits, we define number of rollouts same as the number of reasoning chains finally collected. 
We highlight two key findings: 
\textbf{(1)} With as few as two rollouts, \mits achieves notable improvements in reasoning accuracy, underscoring the effectiveness of PMI-based scoring;
\textbf{(2)} While rStar and CoT-SC exhibit continued performance gains with additional rollouts, RAP occasionally shows performance degradation as the number of rollouts increases. We hypothesize that this degradation stems from RAP's limited action space, which constrains the potential of its MCTS framework.

\paragraph{Effectiveness of different evaluators.}
As we coordinate two models generator \(G\) and evaluator \(E\) for \mits, how different evaluators \(E\) impact the PMI score calculation process are under explored. Therefore, we conduct experiments on the ARC-Challenge dataset using Qwen series models as evaluators, with sizes ranging from 1.5B to 7B, as shown in Table~\ref{tab:arc_generator_evaluator}. 
From the table we can observe that more powerful evaluator models usually yield better results. For example, take \textsc{Qwen2.5-3B} as the generator model, \textsc{Qwen2.5-1.5B} as evaluator can only has accuracy of 86.78\%, while \textsc{Qwen2.5-7B} can have performance of 90.61\%.
This is likely because stronger evaluators provide more accurate probability estimates, leading to more reliable PMI values.

\begin{table}[t]
  \centering

  \begin{minipage}[t]{0.46\linewidth}
    \centering
    \small
    \captionof{table}{Results on ARC-Challenge with different pairs of generator and evaluator models.}
    \label{tab:arc_generator_evaluator}
    \begin{tabular}{lcc}
    \toprule
    & \multicolumn{2}{c}{\textbf{Generator}} \\
    \cmidrule(lr){2-3}
    \textbf{Evaluator} & \textsc{Qwen2.5-3B} & \textsc{\textsc{Qwen2.5-7B}} \\
    \midrule
    \textsc{Qwen2.5-1.5B} &86.78 & 87.20\\
    \textsc{Qwen2.5-3B} & 89.86 & 88.24 \\
    \textsc{Qwen2.5-7B} & \textbf{90.61} & \textbf{91.68} \\
    \bottomrule
\end{tabular}
  \end{minipage}
  \hspace{0.04\linewidth}
  \begin{minipage}[t]{0.46\linewidth}
    \centering
    \small
    \captionof{table}{Results on the StrategyQA dataset with model \textsc{Qwen2.5-3B} and \textsc{Phi-3.5-mini} using different aggregations for PMI scoring.}
    \vspace{-0.11em}
    \label{tab:pmi_sum_avg}
    \begin{tabular}{lcc}
    \toprule
    & \multicolumn{2}{c}{\textbf{Models}} \\
    \cmidrule(lr){2-3}
    \textbf{Aggregation} & \textsc{Qwen2.5-3B} & \textsc{Phi-3.5-mini} \\
    \midrule
    Sum     & 65.89 & 63.26 \\
    Average & \textbf{68.45} & \textbf{66.68} \\
    \bottomrule
\end{tabular}

  \end{minipage}
  \vspace{-0.5em}
\end{table}

\paragraph{Effectiveness of Weighted Average Voting.}
We conduct the ablation study on the Weighted Average Voting technique introduced in Section~\ref{sec:weight_ave}, evaluating the impact of selecting different numbers of top-scoring reasoning paths \(K\) for final answer selection. Using model \textsc{Qwen2.5-3B}, we run \mits with beam width of 32 across all three datasets and vary \(K\) from 1 to 32. The corresponding performance results are reported in Table~\ref{tab:weight_k}.
The results show consistent performance improvements as \(K\) increases, with the steepest gains observed at small \(K\) values (\textit{e.g.}, ARC-Challenge improves from 84.31\% to 89.12\% as \(K\) increases from 1 to 8).
Optimal \(K\) values are dataset-dependent: StrategyQA peaks at \(K=16\) while ARC-Challenge and CommonsenseQA achieve best results at \(K=32\). Performance stabilizes as \(K\) value becomes larger, indicating diminishing returns from incorporating additional reasoning paths.

\begin{table}[ht]
  \centering
  \small
  \setlength{\tabcolsep}{7pt}
  \renewcommand{\arraystretch}{1.15}
  \caption{Performance across different values of top-$K$ for weighted majority voting with \textsc{Qwen2.5-3B}. Best results for each dataset are in \textbf{bold}.}
  \vspace{-0.5em}
  \begin{tabular}{l*{6}{c}}
    \toprule
    \multirow{2}{*}{\textbf{Dataset}} & \multicolumn{6}{c}{\textbf{Top-$K$}} \\
    \cmidrule(lr){2-7}
     & \textbf{1} & \textbf{2} & \textbf{4} & \textbf{8} & \textbf{16} & \textbf{32} \\
    \midrule
    StrategyQA     & 63.34 & 65.41 & 67.44 & 68.04 & \textbf{68.86} & 68.23 \\
    ARC-Challenge  & 84.31 & 85.94 & 87.24 & 89.12 & 89.86 & \textbf{90.68} \\
    CommonsenseQA  & 74.78 & 75.16 & 76.64 & 77.02 & 78.49 & \textbf{78.83} \\
    \bottomrule
  \end{tabular}
  \label{tab:weight_k}
  \vspace{-1em}
\end{table}

\paragraph{Ablation on different PMI variants.}
In section~\ref{sec:exp_setup}, we mention to normalize PMI by averaging its accumulative value over the total steps to mitigate the length bias.
Therefore, we conduct an ablation study comparing these two PMI aggregation approaches: (1) Sum aggregation, which accumulates PMI values across all reasoning steps, and (2) Average aggregation, which divides the cumulative PMI score by the number of steps. Table~\ref{tab:pmi_sum_avg} presents results on the StrategyQA dataset with models \textsc{Qwen2.5-3B} and \textsc{Phi-3.5-mini}.
The results show that The average-aggregated (length-normalized) PMI consistently outperforms the sum-based approach, with improvements of +2.56\% on \textsc{Qwen2.5-3B} and +3.42\% on \textsc{Phi-3.5-mini}. This demonstrates that normalizing by path length prevents certain reasoning paths from being unfairly penalized, enabling fair comparison across paths of different lengths, which promotes the performance simultaneously.

\section{Related Work}
\label{sec:related}

\paragraph{Prompting for LLM Reasoning.} Since the introduction of Chain-of-Thought prompting~\citep{kojima2022large, wei2022chain}, the field has witnessed a rapid development of prompting methods which focus on designing instructions and pipelines to elicit the potential reasoning capabilities within LLMs.
Recent work make progress through several ways, such as decomposing a question into sub-questions to guide the reasoning process~\citep{zhou2022least, khot2023decomposed, yang2023large, zhou2025automating} and demonstration selection~\citep{zhang2023automatic, shum-etal-2023-automatic, diao2024active, shi2025enhancing, yang2025conceptcentric}, which aims to select high-quality exemplars for better prompting performances. 
Higher-order thinking approaches, exemplified by abstract and analogical reasoning~\citep{webb2023emergent, zheng2024take, yasunaga2024large, yu2024thought,liu2025mitigating}, demonstrate effectiveness as well, compared to concrete demonstrations appended in prompts.
Other approaches also include code-integrated prompting~\citep{pmlr-v202-gao23f, chen2023program, Suris_2023_ICCV, pmlr-v235-li24ar, zhou2024solving}, which makes LLMs reason by coding, leading to a more structured way to express logical thinking processes.
These methods aim to improve single-round inference performance and are orthogonal to ours.

\vspace{-0.5em}

\paragraph{Tree Search for LLM Reasoning.} Recent advancements have shown that sampling diverse reasoning paths can significantly enhance performances compared to one-time greedy decoding in both theoretical and empirical ways~\citep{brown2024large, snell2024scaling}, featured by Chain-of-Thought with Self-Consistency (CoT-SC)~\citep{wang2023selfconsistency}. 
Taking one step further, more work guide the reasoning process into tree structures~\citep{yao2023tree, xie2023self}, among which the most pioneering work is Tree-of-Thought~\citep{yao2023tree}. There are also works such as building search tree with beam search~\citep{xie2023self}. Recently, Monte-Carlo Tree Search (MCTS) is regarded a powerful algorithm for advancing LLM reasoning, especially from the perspective of inference-time scaling~\citep{rap, qi2025mutual, ding-etal-2024-everything}.
Moreover, tree structures are also deployed to help LLMs in planning tasks~\citep{hu2024treeplanner, gui2025hypertree}.

\section{Conclusion}
\label{sec:conclusion}

In this work, we proposed Mutual Information Tree Search (\mits), a principled framework that leverages pointwise mutual information (PMI) to guide LLM reasoning without expensive look-ahead simulations. 
\mits introduces three key innovations: PMI-based scoring for reasoning path evaluation, entropy-based dynamic sampling for adaptive compute allocation, and weighted average voting for robust answer selection. 
Through comprehensive experiments across diverse reasoning datasets, \mits substantially outperforms several strong baseline methods by a large margin.
These results demonstrate that information-theoretic principles can effectively steer LLM reasoning processes, achieving superior performances across diverse reasoning tasks while providing a computationally efficient algorithm to traditional tree search methods.
Furthermore, we conduct extensive ablation studies, providing analysis and insights for more advanced LLM reasoning techniques.

\bibliography{iclr2026_conference}
\bibliographystyle{iclr2026_conference}

\clearpage

\appendix
\section{Algorithm Details in \mits}

\subsection{Detailed Derivation for PMI Scoring}
\label{app:pmi_derivation}
In section~\ref{sec:incre_upd}, we obtain the recurrent form for the calculation of PMI scores.
Here in this appendix subsection, we derive and provide the calculation process for PMI score in recurrence relation, starting from Equation~\ref{eq:pmi_def}.

\begin{align}
    \label{pmi}
    \text{PMI}(q; S) & = \log \frac{p(S \mid q)}{p(S)} \notag \\
    & = \log \prod_{i=1}^{n} \frac{p(s_i \mid q, s_1, \dots, s_{i-1})}{p(s_i \mid s_1, \dots, s_{i-1})} \notag \\
    & = \sum_{i=1}^{n} \log \frac{p(s_i \mid q, s_1, \dots, s_{i-1})}{p(s_i \mid s_1, \dots, s_{i-1})} \notag \\
    & = \sum_{i=1}^{n} [\log p(s_i \mid q, s_1, \dots, s_{i-1}) - \log p(s_i \mid s_1, \dots, s_{i-1})]. \notag
\end{align}
Therefore, we have the recurrence relation for PMI score calculation presented in Equation~\ref{eq:pmi_n}.

\subsection{Formula for Dynamic Sampling}
\label{app:dynamic_samp}
In section~\ref{sec:tree}, building upon the entropy-based partitioning described in the main text, we present the detailed mathematical formulation for computing the sampling adjustment \(\Delta N_i\). The proportional control scheme adjusts sampling intensity based on how far the current entropy deviates from the moderate range:
\begin{equation}
\Delta N_i = 
  \begin{cases}
    \min \left(2, \left\lfloor 2 \cdot V_p \cdot \dfrac{H_i - H_{\text{high}}}{H_{\text{high}} - H_{\text{low}}} \right\rceil\right), & H_i > H_{\text{high}}, \\
    0, & H_{\text{low}} \leq H_i \leq H_{\text{high}}, \\
    \max \left(-2, -\left\lfloor 2 \cdot V_p \cdot \dfrac{H_{\text{low}} - H_i}{H_{\text{high}} - H_{\text{low}}} \right\rceil\right), & H_i < H_{\text{low}}.
  \end{cases}
\end{equation}
Here $V_p$ is the proportional gain parameter.
The normalization factor $(H_\text{high} - H_\text{low})$ ensures that the adjustment magnitude is scaled relative to the entropy range, making the controller robust to different entropy scales.
The final number of samples is computed with boundary constraints \(N_i = N_{\text{base}} + \Delta N_i\), where $N_{\text{base}} = 3$ represents the base sample count.
In practice, we enable dynamic sampling only when \(m \ge 10\), ensuring that a sufficiently large number of samples has already been accumulated.

\section{Implementation Details}

\subsection{Hyperparameters}
\label{app:hyperparams}

We conduct a comprehensive hyperparameter search to ensure stable and high-quality performances of the \mits algorithm. Here we provide the key parameters during the inference of LLMs.
Specifically, we mainly record two groups of hyper-parameters, parameters for basic model inference and parameters for search tree construction. As displayed in Table~\ref{tab:hypers}, for critical parameters of basic model inference, we search a set of parameters of temperature, top-k, and top-p for model decoding. 
For search tree construction, the sample number is set to 3, which means by default we sample 3 candidate next-steps given previous steps, as illustrated in section~\ref{sec:method}. The maximum depth of the search tree is 10. For generation of each candidate step, we set the length limit of 512 tokens. For beam width during the \mits process, we set it to 32.

\begin{table}[H]
    \centering
    
    \caption{Hyperparameter Configuration for \mits.}
    \begin{tabular}{ll}
        \toprule
        \textbf{Parameter} & \textbf{Value} \\
        \midrule
        \multicolumn{2}{c}{\textbf{Parameters for Basic Inference}} \\
        
        temperature & 0.9 \\
        top-k & 100 \\
        top-p & 0.96 \\
        
        \addlinespace
        
        \multicolumn{2}{c}{\textbf{Parameters for Search Tree Construction}} \\
        sample numbers & 3 \\
        max depth & 10 \\
        max new tokens & 512 \\
        beam width & 32 \\
        \bottomrule
    \end{tabular}
    
    \label{tab:hypers}
\end{table}

\section{Case Study}
\label{app:case_study}

In this appendix section, we provide some case studies on the datasets StrategyQA, ARC-Challenge, and CommonsenseQA with the backbone model \textsc{Qwen2.5-3B} (both for generator model \(G\) and evaluator model \(E\)). Specifically, we provide the examples in json format and here are the detailed information:
\texttt{"index"} refers to the index of node (reasoning step) in the built search tree, \texttt{"depth"} denotes the depth (level) of the node in the tree, and \texttt{"content"} denotes the content of the reasoning step generated by LLMs. To calculate the PMI score, we record the values of conditional probabilities \(\log p(s_i \mid q, s_1, \dots, s_{i-1})\) and \(\log p(s_i \mid s_1, \dots, s_{i-1})\) of step \(i\), and denote them as \(\log p(S \mid q)\) and \(\log p(S)\) for brevity, respectively. For the PMI score, we denote it as \texttt{"PMI"} and calculate it by simply subtracting two log probabilities \(\log p(S \mid q) - \log p(S)\). In the following subsections, we give case studies one by one for each dataset.

\subsection{StrategyQA}
\paragraph{Prompt.}
Here is an example prompt with question.
\begin{lstlisting}[basicstyle=\small\ttfamily, breaklines=true, breakindent=0em, commentstyle=\color{red!50!green!50!blue!50}, frame=single, rulesepcolor=\color{red!20!green!20!blue!20},numbers=none,literate={`}{\textasciigrave}1]
You need to answer the question with true or false.
Let's think this question step by step. Finally, please make decision from the options A or B. When you believe you've reached the final step, please output your final answer in square brackets like [A. True] or [B. False].
Question:
Is CAS number 8009-03-8 harmful for a rash?
A. True
B. False
Step 1:
\end{lstlisting}

\paragraph{Reasoning Path.}
Here is one of the extracted reasoning paths that correctly solves the problem.
\begin{lstlisting}[basicstyle=\small\ttfamily, breaklines=true, breakindent=0em, commentstyle=\color{red!50!green!50!blue!50}, frame=single, rulesepcolor=\color{red!20!green!20!blue!20},numbers=none,literate={`}{\textasciigrave}1]
[
  {
    "index": 3,
    "depth": 1,
    "content": "Identifying CAS Number. CAS Number 8009-03-8 refers to a compound known as Toluene diisocyanate (TDI).",
    "log p(S|q)": -55.45,
    "log p(S)": -85.48,
    "PMI": 30.04
  },
  {
    "index": 8,
    "depth": 2,
    "content": "Understanding Toluene diisocyanate. Toluene diisocyanate (TDI) is a chemical compound used in the production of polyurethane materials, which can be found in various products like paints, adhesives, and some foam products.",
    "log p(S|q)": -45.46,
    "log p(S)": -54.48,
    "PMI": 9.02
  },
  {
    "index": 23,
    "depth": 3,
    "content": "TDI and Rashes. TDI exposure can cause skin rashes in some individuals, but it is not generally considered to be inherently harmful to the majority of people with rash.",
    "log p(S|q)": -68.64,
    "log p(S)": -106.24,
    "PMI": 37.60
  },
  {
    "index": 42,
    "depth": 4,
    "content": "Conclusion: Based on the information available, TDI is not specifically harmful to rashes for the majority of people. Therefore, the statement that it is harmful for a rash is false. The answer is [B. False].",
    "log p(S|q)": -90.02,
    "log p(S)": -140.89,
    "PMI": 50.87
  }
]
\end{lstlisting}

\subsection{ARC-Challenge}
\paragraph{Prompt.}
Here is an example prompt with question from the dataset ARC-Challenge.
\begin{lstlisting}[basicstyle=\small\ttfamily, breaklines=true, breakindent=0em, commentstyle=\color{red!50!green!50!blue!50}, frame=single, rulesepcolor=\color{red!20!green!20!blue!20},numbers=none,literate={`}{\textasciigrave}1]
You need to answer the question by choosing one of the given options.
Let's think this question step by step. Finally, please make decision from the options. When you believe you've reached the final step, please output your final answer in square brackets like [A. option1], [B. option2], etc.
Question:
A scientist lost the cap to an ammonia bottle, so she moved the ammonia into a clean container with a lid. What else should she do to be safe?
A. Recycle the old ammonia bottle.
B. Label the new container clearly.
C. Heat the new container to seal the lid.
D. Store the new container in the refrigerator.
Step 1:
\end{lstlisting}

\paragraph{Reasoning Path.}
Here is an example path that correctly solves the question.
\begin{lstlisting}[basicstyle=\small\ttfamily, breaklines=true, breakindent=0em, commentstyle=\color{red!50!green!50!blue!50}, frame=single, rulesepcolor=\color{red!20!green!20!blue!20},numbers=none,literate={`}{\textasciigrave}1]
[
  {
    "index": 2,
    "depth": 1,
    "content": "The main issue is the exposure of ammonia, which is a hazardous substance, especially when not properly contained.",
    "log p(S|q)": -57.87,
    "log p(S)": -79.15,
    "PMI": 21.28
  },
  {
    "index": 9,
    "depth": 2,
    "content": "When ammonia is spilled or not properly contained, it can evaporate, leading to inhalation risks and potential health hazards.",
    "log p(S|q)": -42.67,
    "log p(S)": -51.71,
    "PMI": 9.04
  },
  {
    "index": 35,
    "depth": 3,
    "content": "Look at the options. \n- Option A suggests recyling the old ammonia bottle, which is not relevant to the safety concerns of the new container.\n- Option B involves labeling the new container, which is essential. Without proper labeling, others may not realize its contents.\n- Option C suggests heating the new container to seal the lid. This is not effective for sealing a container and could actually cause damage.\n- Option D involves storing the container in the refrigerator. While this might help with the smell, it does not enhance the overall safety.",
    "log p(S|q)": -143.80,
    "log p(S)": -263.64,
    "PMI": 119.84
  },
  {
    "index": 54,
    "depth": 4,
    "content": "Determine the most effective safety measure. Option B ensures that others, including the scientist herself, can identify the contents and take necessary precautions.\n\nTherefore, the most appropriate action to ensure safety is [B. Label the new container clearly].",
    "log p(S|q)": -145.40,
    "log p(S)": -194.64,
    "PMI": 49.24
  }
]
\end{lstlisting}

\subsection{CommonsenseQA}
\paragraph{Prompt.}
Here is an input prompt with example question.
\begin{lstlisting}[basicstyle=\small\ttfamily, breaklines=true, breakindent=0em, commentstyle=\color{red!50!green!50!blue!50}, frame=single, rulesepcolor=\color{red!20!green!20!blue!20},numbers=none,literate={`}{\textasciigrave}1]
You need to answer the question by choosing one of the given options.\nLet's think this question step by step. Finally, please make decision from the options. When you believe you've reached the final step, please output your final answer in square brackets like [A. option1], [B. option2], etc.
Question:
The weasel was becoming a problem, it kept getting into the chicken eggs kept in the what?
A. forrest
B. barn
C. public office
D. out of doors
E. freezer
Step 1:
\end{lstlisting}

\paragraph{Reasoning Path.}
Here is an example path that correctly solves the given question.
\begin{lstlisting}[basicstyle=\small\ttfamily, breaklines=true, breakindent=0em, commentstyle=\color{red!50!green!50!blue!50}, frame=single, rulesepcolor=\color{red!20!green!20!blue!20},numbers=none,literate={`}{\textasciigrave}1]
[
  {
    "index": 3,
    "depth": 1,
    "content": "Context of the Question. The question states that the weasel is causing a problem by gaining access to chicken eggs, which implies a location where eggs are stored and where a weasel could potentially enter.",
    "log p(S|q)": -86.52,
    "log p(S)": -137.24,
    "PMI": 50.72
  },
  {
    "index": 8,
    "depth": 2,
    "content": "Analyze the Options.\n- A. Forrest: Forrests are wild, natural habitats and are not typically used for storing agricultural products.\n- B. Barn: Barns are places where agricultural products, including eggs, are commonly stored.\n- C. Public Office: Public offices are indoor, administrative spaces that are not typically associated with storing agricultural products.\n- D. Out of doors: Out of doors refers to an open, outdoor area, which is not suitable for storing eggs.\n- E. Freezer: A freezer is a cold storage unit designed for keeping food cold, not specifically for storing eggs.",
    "log p(S|q)": -127.25,
    "log p(S)": -210.57,
    "PMI": 83.32
  },
  {
    "index": 16,
    "depth": 3,
    "content": "Determine the Most Likely Answer.\nGiven the context of the question and the function of each option, the most logical place where a weasel would be able to access chicken eggs is inside a barn.",
    "log p(S|q)": -49.20,
    "log p(S)": -68.81,
    "PMI": 19.61
  },
  {
    "index": 43,
    "depth": 4,
    "content": "Therefore, the answer is [B. barn].",
    "log p(S|q)": -66.43,
    "log p(S)": -89.31,
    "PMI": 22.88
  }
]
\end{lstlisting}

% For remark on the normalized PMI and its usage.
% \input{sec/pmi_remark}

\end{document}